%% file: main.tex
\documentclass{article} % For LaTeX2e
\usepackage{iftex}
\ifPDFTeX
  \usepackage[T1]{fontenc}
  \usepackage[utf8]{inputenc}
  \DeclareUnicodeCharacter{0394}{\ensuremath{\Delta}}
\fi
\usepackage{iclr2027_conference,times}

\input{math_commands.tex}

\usepackage[hidelinks]{hyperref}
\usepackage{url}
\usepackage{graphicx}
\usepackage{booktabs}
\usepackage{multirow}
\usepackage{longtable}
\usepackage{placeins}
\usepackage{amsmath,amssymb}
\usepackage{xcolor}
\usepackage{fancyvrb}
\usepackage{fvextra}
\ifPDFTeX
  \DeclareTextSymbol{\textasciigrave}{T1}{0}
  \DeclareTextSymbolDefault{\textasciigrave}{T1}
\fi
\usepackage{tikz}
\usetikzlibrary{positioning,arrows.meta,shapes.geometric,fit,backgrounds,calc}

\graphicspath{{img/asset/}}

\title{Agentic RSR: Real-to-Sim-to-Real through Scene Reconstruction and Execution-Grounded Robot Policies}

\author{
Yihan Li$^{1,\dagger}$ \quad Yating Feng$^2$ \quad Shengjiu Sun$^1$ \quad Jianing Chen$^1$ \quad Hao Ren$^1$\\
Bowen Yang$^2$ \quad Weisheng Xu$^2$ \quad Qiwei Wu$^2$ \quad Hui Cheng$^1$ \quad Renjing Xu$^{2,\dagger}$\\[2pt]
\textsuperscript{1}Sun Yat-sen University\\
\textsuperscript{2}The Hong Kong University of Science and Technology (Guangzhou)\\
\texttt{liyihan.xyz@gmail.com} \quad \texttt{renjingxu@hkust-gz.edu.cn}\\[-1pt]
$^\dagger$ Corresponding authors
}

\iclrfinalcopy

\begin{document}

\maketitle
\lhead{Preprint}

\input{section/00_abstract}
\input{img/fig_teaser}
\input{section/01_introduction}
\input{section/02_related_work}
\input{tab/system_comparison}
\input{section/03_method}
\input{section/04_experiments}
\input{section/05_conclusion}
\input{section/06_ai_use}

\bibliography{ref}
\bibliographystyle{iclr2027_conference}

\clearpage
\appendix
\input{appendix/task_catalog}
\input{appendix/stage1}
\input{appendix/stage2}
\input{appendix/agent_prompt_templates}

\end{document}

%% file: math_commands.tex
\usepackage{amsmath,amsfonts,bm}

\def\eqref#1{equation~\ref{#1}}
\def\1{\bm{1}}

\DeclareMathAlphabet{\mathsfit}{\encodingdefault}{\sfdefault}{m}{sl}
\SetMathAlphabet{\mathsfit}{bold}{\encodingdefault}{\sfdefault}{bx}{n}

%% file: section/00_abstract.tex
\begin{abstract}
A simulation of a real robot workspace must preserve task-relevant interactions, while policies developed in it must operate on observations available to the real robot. Yet scene reconstruction and policy development are often treated separately. We present \textbf{Agentic Real-to-Sim-to-Real (Agentic RSR)}, a framework that links scene reconstruction, policy development, and real-robot execution through the same manipulation task. Given a workspace video, a task description, and a known robot model, an agent recovers metric scale, iteratively refines the scene using visual feedback, and checks task-relevant interactions in MuJoCo. A coding agent then develops an executable policy, progressing from privileged object poses to visual observations and randomized simulation. The policy can interleave multiple observations and actions within one invocation, while the agent uses execution feedback to continue, retry, or revise its approach. A shared task-level interface carries the policy and accumulated experience to the real robot, where fresh observations and safety checks guide execution. Using Claude Opus 5.5 through DeepSeek Harness (DSH), all 18 scenes across two robots pass the joint visual check; mean four-view Depth MAE against reference depth estimates is 0.0808~m, mean Lab $\Delta E_{76}$ is 8.01, and mean grayscale SSIM is 0.7338. In real-robot experiments, the aggregate task success rate reaches 80\% of the simulation task success rate, indicating substantial retention of simulated performance on hardware. Code and reconstructed scene data will be made publicly available.
\end{abstract}

%% file: img/fig_teaser.tex
\begin{figure}[!ht]
    \centering
    \includegraphics[width=0.9\linewidth]{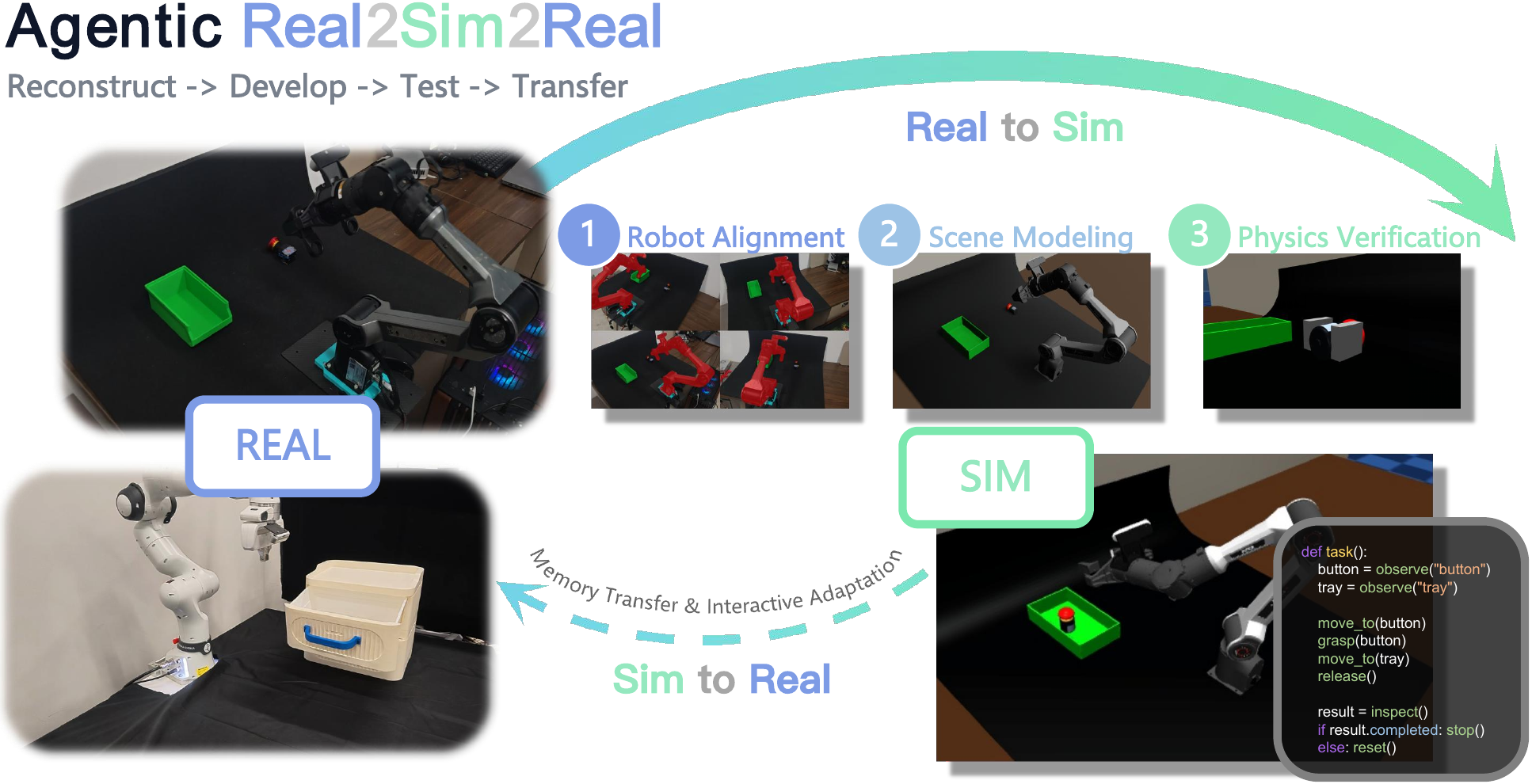}
    \caption{\textbf{Agentic RSR overview.} An agent uses real observations and known robot geometry to reconstruct a metrically aligned scene and check task-relevant interactions. A policy agent develops and tests an executable policy in simulation, then transfers it with experience memory to the real robot for feedback-driven adaptation.}
    \label{fig:teaser}
\end{figure}

%% file: section/01_introduction.tex
\section{Introduction}
\label{sec:intro}

Developing and evaluating robot policies through real-world trials is costly and difficult to scale. Simulation offers an alternative for policy evaluation and training~\citep{li2024simpler,ranawaka2026simfoundry}. For a policy intended to return to a captured workspace, the simulated scene must support task-relevant interactions, and the policy must operate on observations available to the real robot.

Existing systems already bridge parts of this chain. Re$^3$Sim reconstructs interactive scenes and demonstrates sim-to-real policy transfer~\citep{han2025re3sim}; Agentic Real2Sim uses vision-language agents to build physics-based episodic scenes for downstream policy evaluation and fine-tuning~\citep{chen2026agentic}; and SimFoundry turns video into interactive scenes for simulation-based evaluation and sim-to-real policy training~\citep{ranawaka2026simfoundry}. In programmatic control, Code as Policies generates executable programs over perception and control APIs~\citep{liang2023code}, while CaP-X studies multi-turn execution feedback and transfers code-based policies using tools shared across simulation and reality~\citep{capx2026}. This motivates a more specific question: how can agent-refined reconstruction of an observed workspace be coupled with an execution-grounded programmatic policy and a consistent task-level interface across simulation and hardware?

We present \textbf{Agentic Real-to-Sim-to-Real (Agentic RSR)} to study this connection. Given a workspace video, a task, and a known robot, a DeepSeek Harness (DSH) agent reconstructs a metrically aligned, interactive scene and refines it against visual and task-relevant physical checks. A Codex agent~\citep{openai2026codexsdk} then develops an executable programmatic policy, progressing from privileged object poses to visual observations and randomized simulation. The policy may be revised during development, but its version is locked during final evaluation. A shared task-level interface carries the policy and experience memory to hardware, where fresh observations and execution feedback support verification and recovery. Real-robot experiments show that the policies retain much of their aggregate task performance after transfer from simulation.

Our contributions are:
\begin{itemize}
    \item \textbf{Agentic Real-to-Sim scene reconstruction for executable tasks.} We anchor the metric scale of a video-reconstructed workspace to known robot geometry, use DSH agents to iteratively reconstruct the scene using reference views and depth and appearance feedback, and check task-relevant geometric and physical interactions so that the scene supports robot execution;
    \item \textbf{An execution-grounded programmatic policy developed by a Codex agent.} The Codex agent~\citep{openai2026codexsdk} generates a policy bundle containing a complete program and reusable steps. The bundle remains locked during an episode, while the Codex agent~\citep{openai2026codexsdk} uses execution results and fresh observations to decide whether to continue, switch steps, retry, or recover;
    \item \textbf{A unified Agentic Real-to-Sim-to-Real closed-loop framework.} A consistent task-level observation and action interface connects scene reconstruction, simulation-based policy testing, and hardware execution, where execution feedback supports verification and recovery.
\end{itemize}

%% file: section/02_related_work.tex
\section{Related Work}
\label{sec:related}

\paragraph{\textbf{Real-to-Sim Scene Reconstruction for Robot Policies.}} RL-GSBridge and GSWorld combine high-fidelity scene representations with robot simulation, while Re$^3$Sim and RoboSimGS couple rendering with interaction geometry for manipulation and sim-to-real learning~\citep{wu2024rlgsbridge,jiang2025gsworld,han2025re3sim,zhao2025robosimgs}. RoboSnap builds simulation-ready scenes from a single RGB image~\citep{zhang2026robosnap}. Scalable Real2Sim recovers physics-aware assets through robot interaction, whereas TwinAligner addresses visual and dynamic alignment for policy transfer~\citep{pfaff2025scalable,fan2025twinaligner}.

ACDC generates \emph{digital cousins} that preserve task-relevant affordances, while SIMPLER studies how simulated evaluation relates to real-robot performance~\citep{dai2024acdc,li2024simpler}. Holodeck generates prompt-conditioned environments; Gen2Sim produces 3D assets, tasks, and rewards for robot learning~\citep{yang2024holodeck,katara2024gen2sim}. SceneSmith, PhyScensis, SimWorld Studio, and Articraft respectively explore vision-language scene generation, physics-guided arrangement, coding-agent environments, and programmatic articulated assets~\citep{pfaff2026scenesmith,wang2026physcensis,kang2026simworld,zhou2026articraft}. Closer to video-based real-to-sim, Agentic Real2Sim uses agents and simulator-in-the-loop refinement to turn recorded robot--object interactions into simulatable episodic twins, while SimFoundry generates interactive scenes and \emph{digital cousins} for policy evaluation and sim-to-real training~\citep{chen2026agentic,ranawaka2026simfoundry}.

\paragraph{\textbf{Programmatic Robot Policies and Execution-Grounded Agents.}} Code as Policies and Instruct2Act compose perception and actions into executable programs~\citep{liang2023code,huang2023instruct2act}. SayCan pairs language planning with feasible pretrained skills, while VoxPoser uses 3D value maps for closed-loop trajectory planning~\citep{ahn2022saycan,huang2023voxposer}. Inner Monologue incorporates environmental feedback into planning; CaP-X studies multi-turn coding agents and code-policy transfer across simulation and hardware~\citep{huang2022inner,capx2026}. ASPIRE and ENPIRE explore feedback-driven skill development; Show-Harness offers a semantic action interface, and RoboClaw orchestrates learned primitives for long-horizon tasks~\citep{lu2026aspire,xiao2026enpire,chen2026showharness,li2026roboclaw}.

Visual and dynamics randomization support sim-to-real transfer~\citep{tobin2017domain,peng2018simtoreal}, while RT-2, OpenVLA, Octo, and $\pi_0$ develop data-driven visuomotor rather than executable robot policies~\citep{brohan2023rt2,kim2024openvla,geng2024octo,black2024pi0}. Our focus is to connect reconstruction of a particular captured workspace with a programmatic policy for the same task, evaluated in simulation and on real robots.

%% file: tab/system_comparison.tex
\begin{table}[htbp]
\caption{\textbf{System-level capability comparison.} Reported capabilities in real-scene reconstruction, agentic scene revision, code-policy generation, execution-time recovery, and sim-to-real trials. Symbols are defined below the table.}
\label{tab:system-comparison}
\centering
\scriptsize
\setlength{\tabcolsep}{2pt}
\renewcommand{\arraystretch}{1.15}
\begin{tabular*}{\linewidth}{@{\extracolsep{\fill}}p{0.32\linewidth}*{6}{c}@{}}
\toprule
Capability & \rotatebox[origin=c]{55}{CaP} & \rotatebox[origin=c]{55}{Agentic R2S} & \rotatebox[origin=c]{55}{CaP-X} & \rotatebox[origin=c]{55}{Re$^3$Sim} & \rotatebox[origin=c]{55}{SimFoundry} & \rotatebox[origin=c]{55}{\shortstack{Agentic RSR\\(Ours)}} \\
\midrule
Real-scene reconstruction & $\times$ & \checkmark & $\times$ & \checkmark & \checkmark & \checkmark \\
Agentic scene revision & $\times$ & \checkmark & $\times$ & $\times$ & $\times$ & \checkmark \\
Code-policy generation & \checkmark & $\times$ & \checkmark & $\times$ & $\times$ & \checkmark \\
Runtime agent recovery & -- & $\times$ & \checkmark & $\times$ & $\times$ & \checkmark \\
Sim-to-real trials & -- & -- & \checkmark & \checkmark & \checkmark & \checkmark \\
\bottomrule
\end{tabular*}
\par\smallskip{\raggedright\scriptsize \checkmark: experimentally demonstrated; --: not established in the paper; $\times$: not part of the implemented method.\par}
\end{table}

%% file: section/03_method.tex
\section{Method}
\label{sec:method}

\subsection{Real-to-Sim scene reconstruction for executable tasks}
\label{sec:method:real2sim}

\paragraph{\textbf{Task and scene output.}} Given a continuous RGB video $V$ of a workspace, a task description $q$, and a known robot asset $U$, we aim to reconstruct the particular captured workspace rather than generate a semantically similar scene variant. The resulting scene should reproduce task-relevant geometry and appearance from reference views while providing collision and physical representations that support robot--object interaction. The workflow connects robot-anchored metric registration, feedback-guided agentic scene reconstruction, and task-interaction checks in MuJoCo~\citep{todorov2012mujoco}. The robot asset $U$ supplies both a metric geometric reference and the kinematic model used in downstream simulation.

\input{img/agentic_rsr}

\paragraph{\textbf{Robot-anchored metric registration.}} We sample 32 frames from $V$ and select four as fixed reference views for subsequent scene reconstruction. VGGT-$\Omega$~\citep{wang2026vggtomega} estimates camera intrinsics $K_v$, world-to-camera extrinsics $E_v$, and depth $d_v$ for sampled frame $v$, while SAM3~\citep{carion2025sam3} supplies robot masks. Back-projecting masked depths yields a robot point cloud in the reconstruction frame. We align the known meter-scale URDF surface with this cloud using multi-scale Sim(3) registration, followed by local pose and scale refinement. Let $T_{B\rightarrow G}$ denote the similarity transform from robot-base frame $B$ to reconstruction frame $G$, with scale $s$ measured in reconstruction units per meter. Applying $T_{B\rightarrow G}^{-1}$ expresses the reconstructed scene and camera poses in the metric robot-base frame; dividing the estimated depths by $s$ converts them to meters while preserving their relative geometry. Candidate registrations are ranked by projected-mask agreement, bidirectional point-cloud fit, coverage, and depth consistency. Appendix~\ref{app:scene-setup} specifies the objective and search settings.

\paragraph{\textbf{Feedback-guided agentic scene reconstruction.}} We use DeepSeek Harness (DSH) to orchestrate separate agent sessions for scene planning, refinement, and physical checks. DSH is the execution framework, distinct from the model selected for each session; our main reconstruction configuration uses Claude Opus 5.5. A planning session first receives $q$ and four real reference RGB images and produces a plan for task objects, background elements, and spatial relationships. A separate refinement session receives this plan, the initial Blender~\citep{blender2026} scene, the fixed reference-camera status, the real images, and RGB previews of the current scene. Across successive rounds, it edits object geometry, relative layout, materials, and lighting. The robot and four reference cameras are frozen by the execution program and cannot be changed by the agent. The agent can use Blender~\citep{blender2026} Lab MCP tools for scene editing and preview when available. In each round, the refinement agent renders and inspects the candidate before submitting a Blender~\citep{blender2026} Python edit script. An external runner applies the script, checks the frozen assets, saves a separate candidate version, and renders and evaluates it from the four reference views. The agent receives scalar feedback for overall depth and appearance error; acceptance is determined by the programmatic evaluator rather than the agent's self-assessment.

For reference view $v$, the evaluator records metric depth error $D_v$, Lab color difference $C_v$, and valid rendered-depth coverage $Q_v$. Equal-weight averages over the four views give $\overline D$ and $\overline C$. The joint visual check requires $\overline D<0.15$~m and $\overline C<15$, together with $D_v<0.20$~m, $C_v<30$, and $Q_v\geq0.50$ for every view. Otherwise, refinement continues or records failure at the iteration limit. Reaching the mean-depth threshold alone does not establish full acceptance. Since the reference depths come from the preceding video reconstruction, $D_v$ measures consistency with that metric reference rather than geometric error against independently measured ground truth. DSH aggregate results appear in Table~\ref{tab:dsh-models}; historical Codex per-scene results appear in Appendix~\ref{app:quantitative-results}.

\paragraph{\textbf{MuJoCo scene and task-interaction checks.}} For a candidate that passes the visual check, a separate DSH agent session receives the task description, scene plan, mesh inventory, and reference-view images. It classifies meshes as visual-only elements, static support or collision geometry, or movable or articulated task objects. These classifications are reviewable physical hypotheses, not measurements of real object properties. An exporter then combines the scene with the known robot model to generate a MuJoCo~\citep{todorov2012mujoco} scene $\mathcal S_M$ with appropriate collision representations and simulated physical properties. A separate DSH physics-checking session can use repository-provided inspection tools to examine task-object states and run settling, simplified grasp, or articulated-motion probes. Where justified, it tests a separate candidate with changes to collision geometry, object pose, or physical parameters. The runner records tool calls and reports and reruns the relevant probes on revised candidates; visible geometry or pose changes also trigger regression checks from the fixed reference views.

\input{img/sim_real}
\FloatBarrier

\subsection{Simulation-based policy generation and execution}
\label{sec:method:sim}

\paragraph{\textbf{Three-pass policy iteration.}} Policy development proceeds through three successive passes. First, ground-truth object poses are temporarily available in the nominal scene, allowing the agent to establish task decomposition and action logic. Second, this privileged access is removed: the policy must estimate object state through the visual observation interface intended for subsequent real-robot use and complete the task in the nominal scene. Third, the same visual interface is retained while bounded variation is introduced in initial object positions, camera pose and calibration, and action execution. This pass tests whether the policy adapts to new observations and execution outcomes. The agent may revise its policy within each pass; when it judges the task complete from observations, it renders a final frame as evidence before passing the policy and experience notes to a fresh agent conversation for the next pass. These passes constitute iterative development; the submitted policy version is held fixed during final evaluation, while its actions can still respond to observations.

\paragraph{\textbf{Policy-facing interface.}} The simulator loads a reconstructed MuJoCo~\citep{todorov2012mujoco} scene and provides a consistent task-level interface for Piper and Franka. State and observation interfaces provide the end-effector state, estimated wrist-camera pose, gripper state, wrist and third-person images, and object-position estimates derived from wrist RGB-D and SAM3~\citep{carion2025sam3}. Action interfaces support end-effector pose commands, gripper control, and sparse trajectories; control-flow interfaces support scene reset, yielding control, and stopping execution. Apart from the temporary object-pose access in the first pass, the visual policy cannot read simulator truth or rely on a built-in pick-and-place action or task-success oracle.

\input{img/policy_iteration_loop}

\paragraph{\textbf{Execution-grounded policy loop.}} In each development pass, the Codex agent~\citep{openai2026codexsdk} writes a programmatic policy using the task description and public interface. A single program execution may interleave repeated observations and actions. If execution fails or yields control, the runner returns action outcomes, errors, and recent observations to the same agent conversation. The agent can then continue from the current scene, reset and retry, or revise the policy before another execution. Yielding control preserves the scene state; resetting starts a new attempt and resamples perturbations in the randomized pass. The agent decides whether the task is complete from its observations and provides a final rendered frame as evidence; it does not receive a built-in task-success oracle.

\subsection{Sim-to-real transfer of policy and experience memory}
\label{sec:method:real}

\paragraph{\textbf{Policy and experience memory.}} The simulation phase transfers two complementary forms of information: the policy retains validated task steps and observation--action logic, while experience memory summarizes effective behaviors, failure causes, perception difficulties, and actions to try or avoid. This memory helps the agent adapt to the real environment; it does not provide ground-truth object poses or simulator internals. A new agent conversation begins with the policy and experience memory but must re-localize objects and assess task state from current real observations rather than reuse simulated coordinates or success claims.

\paragraph{\textbf{Robot-side safety constraints and online verification.}} During transfer, a robot-side safety mechanism automatically checks coordinate frames and camera calibration, enforces action limits and a safe workspace specific to the robot and task, and gates execution accordingly. Actions that violate these constraints are not executed. In parallel, the real-robot policy instructions require the agent to observe more frequently within a single program invocation, particularly before and after consequential actions, so that it reconfirms object locations and robot state before continuing, adjusting, or stopping. Real execution outcomes may update the experience memory; the agent assesses task completion from current observations rather than relying on a built-in success signal.

%% file: img/agentic_rsr.tex
\begin{figure}[!ht]
    \centering
    \includegraphics[width=\linewidth]{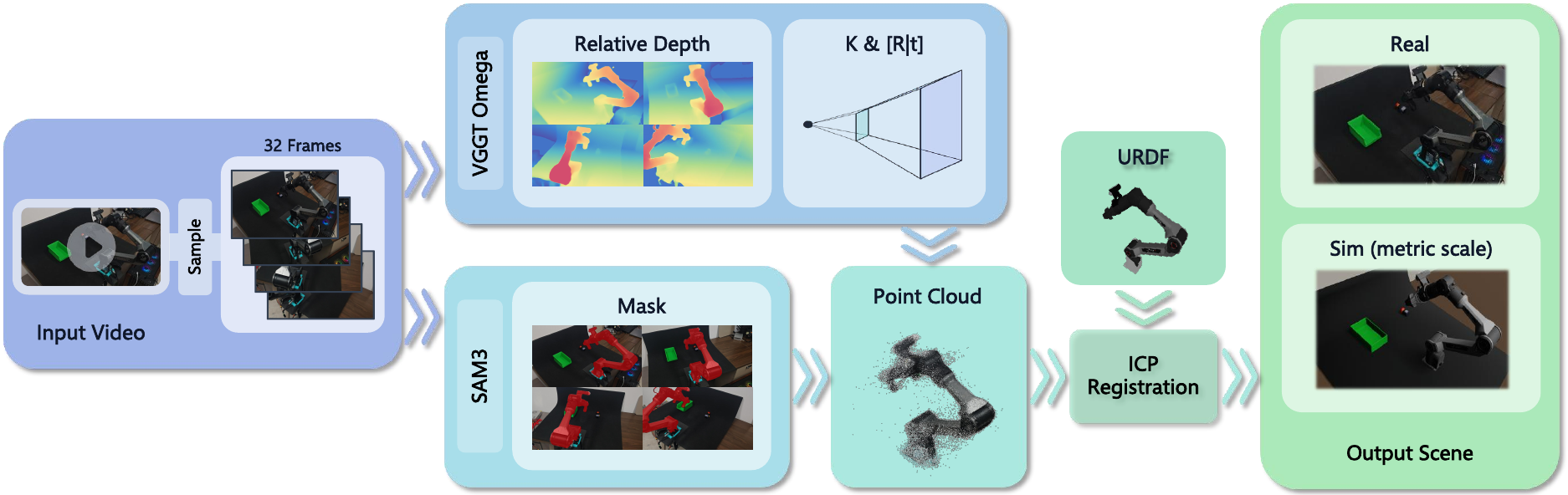}
    \caption{\textbf{Recovering metric scale from known robot geometry.} Video-estimated depth, camera parameters, and robot masks are used to construct an observed robot point cloud. Aligning this point cloud with the known URDF geometry yields metric depth and camera poses expressed in the robot-base frame.}
    \label{fig:real2sim-pipeline}
\end{figure}

%% file: img/sim_real.tex
\begin{figure}[!ht]
    \centering
    \includegraphics[width=\linewidth]{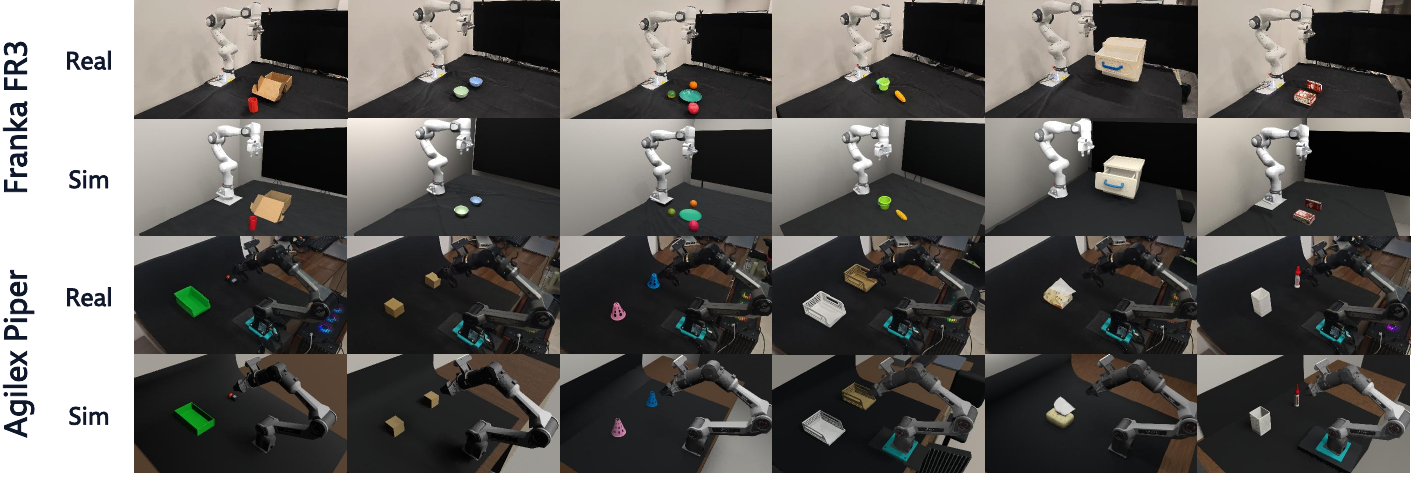}
    \caption{\textbf{Real and simulated views of a subset of task scenes.}}
    \label{fig:real-sim-comparison}
\end{figure}

%% file: img/policy_iteration_loop.tex
\begin{figure}[!ht]
    \centering
    \includegraphics[width=\linewidth]{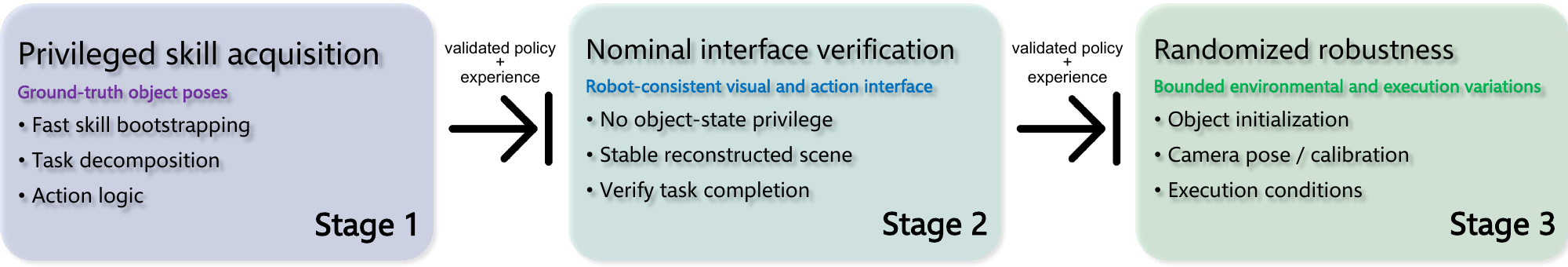}
    \caption{\textbf{Three-stage policy development.} The policy progresses from privileged skill acquisition to nominal interface verification and randomized robustness, passing validated policy and experience between stages.}
    \label{fig:policy-stages}
\end{figure}

%% file: section/04_experiments.tex
\section{Experiments}
\label{sec:exp}

We evaluate Agentic RSR on 18 real-world manipulation scenes and tasks across two robots, covering DSH-based scene reconstruction, model comparisons, and simulated and real-robot policy execution. Historical Codex reconstruction and component ablations are documented separately in the appendix.

\subsection{Experimental Setup}
\label{sec:exp:setup}

Each of the 18 tasks is paired with a workspace RGB video. All scene-reconstruction methods receive the same source video and select frames according to their input requirements; method-specific additional capture is reported separately. The camera slowly circles the workspace, with each frame showing the robot, scene, and task objects. Real-to-Sim planning, refinement, and physical checks use DeepSeek Harness (DSH), with Claude Opus 5.5 as the main configuration. The DSH runs retain the same 18 scenes, four reference views, robot-registration inputs, and metric definitions used for the earlier reconstruction evaluation. Simulation and real-robot policy agents retain the Codex implementation~\citep{openai2026codexsdk} powered by GPT-6 Astra with high reasoning effort.

We measure scene quality with depth mean absolute error (Depth MAE), CIE 1976 Lab color difference (Lab $\Delta E_{76}$), and structural similarity index measure (SSIM). When object-level geometry is available, we also report symmetric Chamfer distance (CD), F1 at a 1~cm threshold (F1@1cm), and bounding-box center error.

For policy evaluation, task success rate (SR) is the fraction of completed tasks for Instruct2Act~\citep{huang2023instruct2act} and Agentic RSR; Code as Policies~\citep{liang2023code} instead reports a task-equal average of recorded real-robot attempts. We report simulation success rate (Sim SR) and real-robot success rate (Real SR). For Code as Policies and Instruct2Act, conversion is the descriptive ratio $100\times\mathrm{Real\ SR}/\mathrm{Sim\ SR}$ from unrounded rates; it is undefined when Sim SR is zero and is not paired task transfer. For Agentic RSR, conversion is the fraction of simulation-successful tasks that also succeed on the real robot. All aggregate results weight scenes or tasks equally.

\input{tab/scene_baselines}
\FloatBarrier

\subsection{Real-to-Sim Scene Reconstruction}
\label{sec:exp:scene}

We first test whether workspace videos can be converted into metric, visually consistent, task-relevant simulation scenes. Robot segmentation and registration use VGGT-$\Omega$~\citep{wang2026vggtomega}, SAM3~\citep{carion2025sam3}, and the known robot URDF to recover scale and robot--camera alignment. Scene planning and refinement construct the task scene in Blender~\citep{blender2026} while keeping the registered robot and reference cameras fixed; the selected scene is converted to MuJoCo~\citep{todorov2012mujoco} for interaction.

With Claude Opus 5.5 through DSH, all 18 scenes pass the joint visual check within the five-round budget, with a mean first acceptance round of 2.22. Mean four-view Depth MAE is 0.08078~m, mean Lab $\Delta E_{76}$ is 8.0104, and mean full-frame grayscale SSIM is 0.73375. Acceptance uses the complete depth, color, and coverage conditions in Section~\ref{sec:method:real2sim}; reaching the mean-depth threshold alone does not establish acceptance. Object-level geometry metrics have not been reported for this DSH batch. The earlier Codex reconstruction metrics, optimization-round statistics, and object-geometry evaluation are retained in Appendix~\ref{app:codex-history}.

We additionally evaluate Re$^3$Sim~\citep{han2025re3sim}, RL-GSBridge~\citep{wu2024rlgsbridge}, RoboSimGS~\citep{zhao2025robosimgs}, and SimFoundry~\citep{ranawaka2026simfoundry} on the same 18 task videos. These baseline results are retained from the earlier evaluation. All methods share the source videos, but differ in whether they process video directly or selected multi-view frames. Missing baseline entries in Table~\ref{tab:scene-baselines} reflect output incompatibility: some methods produce images or Gaussian representations without depth or instance-level geometry directly usable here, not an inherent inability of Gaussian representations to produce depth. Among the reconstruction methods in this table, the DSH configuration has the lowest reported depth and color errors and the highest SSIM. Historical per-scene comparisons appear in Appendix~\ref{app:re3sim-audit}.

\subsection{Model Evaluation with DSH}
\label{sec:exp:models}

Table~\ref{tab:dsh-models} reports the eight DSH configurations in the October 6, 2026 evaluation summary. Every configuration is attempted on all 18 scenes. For the six configurations with scene metrics, $N_{\mathrm{metric}}=18$: means include all scenes, including those that do not pass. $R_{\mathrm{pass}}$ averages the first round satisfying the complete joint visual check over passing scenes only; it is distinct from the depth-only first-hit statistic $R_{0.15}$ used in the historical Codex ablations.

Claude Opus 5.5, the main configuration, and Claude Fable 5.1 both pass 18/18 scenes. Fable has the lowest mean Depth MAE (0.07326~m), while Opus has the lowest mean Lab $\Delta E_{76}$ (8.0104) and the highest mean SSIM (0.73375) among the DSH configurations. GPT-6 Astra and Claude Sonnet 5.5 each pass 15/18 scenes, GPT-6.1 Sol passes 12/18, and GPT-5.6 Sol passes 4/18. These outcomes describe performance within the evaluated workflow rather than a general model ranking.

Kimi K3 and DeepSeek V4.1 Flash each pass 0/18 scenes: in these runs, neither model completes the tool interactions required to execute the workflow. No valid scene metrics are available, so $N_{\mathrm{metric}}=0$ and metric entries are --; these entries do not represent zero scene error. DSH names the harness and is distinct from the DeepSeek model evaluated here.

The three GPT DSH configurations, Claude Opus 5.5, and Claude Sonnet 5.5 use a five-round limit. The Fable round limit is not reported in the supplied summary; round budgets for the two tool-interaction failures are also unreported. The latest Codex summaries and the earlier component ablations are retained in Appendix~\ref{app:codex-history}. Because the harness was refactored and iteration budgets differ or are unreported, cross-batch comparisons are descriptive and do not isolate model or harness effects.

\input{tab/dsh_models}
\FloatBarrier

\subsection{Simulation and Real-Robot Policy Evaluation}
\label{sec:exp:sim}

We report simulation and real-robot results for Code as Policies~\citep{liang2023code}, Instruct2Act~\citep{huang2023instruct2act}, and Agentic RSR across the 18 tasks, retaining each method's native policy format.

We measure success under randomized simulation conditions and recorded real-robot attempts. Table~\ref{tab:task-comparison} reports aggregate results; task IDs are listed in Appendix~\ref{app:task-catalog}.

Code as Policies~\citep{liang2023code} reports Sim SR and task-equal real-robot success of 50.0\% and 25.6\%, respectively, for a descriptive conversion ratio of 51.1\%. Instruct2Act~\citep{huang2023instruct2act} succeeds on 7/18 simulation tasks and 5/18 real-robot tasks, yielding Sim SR, Real SR, and conversion of 38.9\%, 27.8\%, and 71.4\%.

Agentic RSR completes 10/18 tasks in simulation, 8 of which also succeed on the real robot, yielding Sim SR, Real SR, and paired transfer of 55.6\%, 44.4\%, and 80.0\%.

\input{img/real_robot_execution}
\FloatBarrier

Perception and manipulation technique both affect execution. A single wrist-camera view can be occluded or too narrow, and baseline records include cases where task objects were not recognized. Agentic RSR instead iterates on task steps in simulation and uses execution feedback to reobserve, adjust, or retry, responding to viewpoint changes and post-manipulation object states.

For example, bowl stacking requires stable grasp and placement contacts, not just object localization. Repeated simulation attempts let the agent encode such techniques in policy steps and experience for real-robot execution. Table~\ref{tab:task-comparison} therefore captures revision from execution outcomes, not only one-shot action generation. Because Code as Policies~\citep{liang2023code} hardware scenarios do not fully match the indexed tasks, real-robot comparisons remain descriptive rather than paired.

\input{tab/task_comparison}
\FloatBarrier

%% file: tab/scene_baselines.tex
\begin{table}[!ht]
\caption{Scene reconstruction results across 18 tasks. Agentic RSR uses Claude Opus 5.5 through DSH. -- indicates an unreported metric; object-level geometry metrics have not been reported for this DSH batch.}
\label{tab:scene-baselines}
\centering\scriptsize
\setlength{\tabcolsep}{2pt}
\begin{tabular}{@{}llcccccc@{}}
\toprule
Method & Input & \shortstack{Depth MAE\\(m)$\downarrow$} & $\Delta E_{76}\downarrow$ & SSIM $\uparrow$ & CD (m)$\downarrow$ & F1@1cm$\uparrow$ & BBox (m)$\downarrow$ \\
\midrule
Re$^3$Sim~\citep{han2025re3sim} & Multi-view & -- & 20.60 & 0.5835 & -- & -- & -- \\
RL-GSBridge~\citep{wu2024rlgsbridge} & Video & -- & 22.87 & 0.5568 & -- & -- & -- \\
RoboSimGS~\citep{zhao2025robosimgs} & Multi-view & -- & 20.43 & 0.5939 & -- & -- & -- \\
SimFoundry~\citep{ranawaka2026simfoundry} & Video & 0.3916 & 60.48 & 0.4495 & 0.1736 & 0.0084 & 0.2355 \\
Agentic RSR (Ours) & Video & 0.08078 & 8.0104 & 0.73375 & -- & -- & -- \\
\bottomrule
\end{tabular}
\end{table}

%% file: tab/dsh_models.tex
\begin{table}[!ht]
\caption{DSH scene-refinement results on 18 scenes per configuration. Best in bold. $N_{\mathrm{metric}}$ is the metric denominator; $R_{\mathrm{pass}}$ averages first joint acceptance over passing scenes only. -- denotes unavailable results.}
\label{tab:dsh-models}
\centering\scriptsize
\setlength{\tabcolsep}{2pt}
\begin{tabular*}{\linewidth}{@{\extracolsep{\fill}}lcccccc@{}}
\toprule
Model & $N_{\mathrm{metric}}$ & Pass & $R_{\mathrm{pass}}$ & \shortstack{MAE (m)\\$\downarrow$} & $\Delta E_{76}\downarrow$ & SSIM $\uparrow$ \\
\midrule
\input{tab/dsh_model_rows}
\end{tabular*}
\end{table}

%% file: tab/dsh_model_rows.tex
GPT-6 Astra & 18 & 15/18 & 2.73 & 0.11649 & 10.8432 & 0.69389 \\
GPT-5.6 Sol & 18 & 4/18 & 2.50 & 0.23073 & 18.2622 & 0.60854 \\
GPT-6.1 Sol & 18 & 12/18 & 2.50 & 0.14766 & 11.5203 & 0.68940 \\
Claude Fable 5.1 & 18 & \textbf{18/18} & 2.50 & \textbf{0.07326} & 11.0526 & 0.67221 \\
Claude Opus 5.5 & 18 & \textbf{18/18} & \textbf{2.22} & 0.08078 & \textbf{8.0104} & \textbf{0.73375} \\
Claude Sonnet 5.5 & 18 & 15/18 & 3.40 & 0.12848 & 12.4217 & 0.68008 \\
Kimi K3 & 0 & 0/18 & -- & -- & -- & -- \\
DeepSeek V4.1 Flash & 0 & 0/18 & -- & -- & -- & -- \\
\bottomrule

%% file: img/real_robot_execution.tex
\begin{figure}[!htbp]
    \centering
    \includegraphics[width=\linewidth]{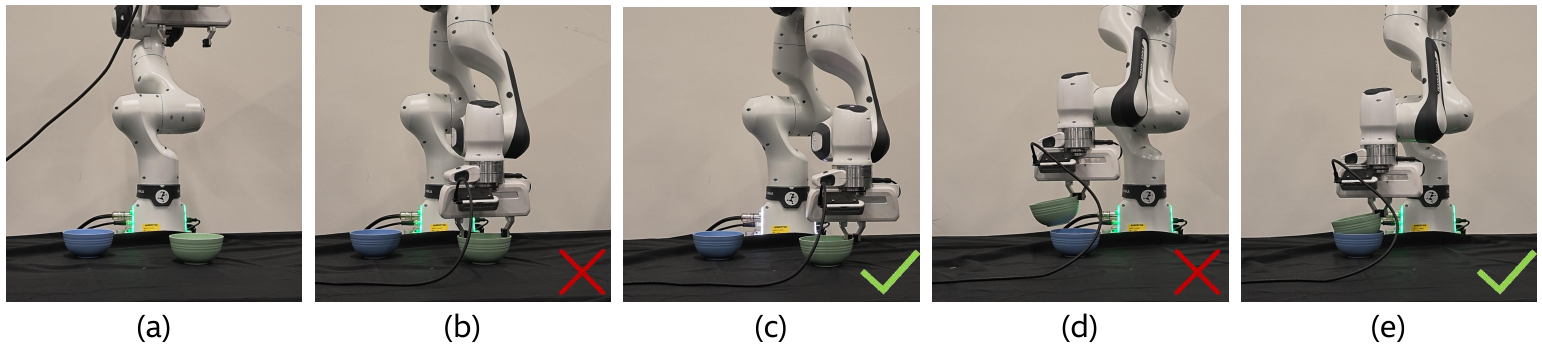}
    \caption{\textbf{Representative real-robot execution frames.} The initial scene is shown in (a); red crosses and green checks mark failed and successful placement outcomes in the subsequent trials.}
    \label{fig:real-robot-execution}
\end{figure}

%% file: tab/task_comparison.tex
\begin{table}[!ht]
\caption{Simulation and real-robot policy results across 18 tasks. For the Code as Policies and Instruct2Act baselines, conversion is the ratio of reported real-robot to simulation success rates, not paired task transfer. For Agentic RSR, conversion measures paired transfer among tasks that succeed in simulation.}
\label{tab:task-comparison}
\centering\footnotesize
\begin{tabular}{@{}lccc@{}}
\toprule
Method & Sim SR (\%) $\uparrow$ & Real SR (\%) $\uparrow$ & Conversion (\%) $\uparrow$ \\
\midrule
Code as Policies~\citep{liang2023code} & 50.0 & 25.6 & 51.1 \\
Instruct2Act~\citep{huang2023instruct2act} & 38.9 & 27.8 & 71.4 \\
Agentic RSR (Ours) & \textbf{55.6} & \textbf{44.4} & \textbf{80.0} \\
\bottomrule
\end{tabular}
\end{table}

%% file: section/05_conclusion.tex
\section{Limitations}
\label{sec:limitations}

In our earlier exploratory use of GPT-6 Astra with the Codex implementation, we observed noticeable differences in agent behavior across devices and invocation interfaces, as well as between runs on the same device at different times. These observations have not been measured under a controlled, matched-input protocol, so their causes remain unclear. Such variability makes agent behavior difficult to reproduce when relying on a closed-source model service. We hope for stronger transparency and oversight of these services, including clear model versioning and stable invocation interfaces, so that researchers can access them consistently and reasonably reproduce experimental results.

Programmatic control and agent-driven revision present different trade-offs between execution speed and adaptability. Code as Policies (CaP) turns task instructions into executable programs that can themselves contain perception-feedback loops~\citep{liang2023code}. Instruct2Act likewise generates programs that compose perception, planning, and robot actions from multimodal instructions~\citep{huang2023instruct2act}. When execution encounters a case outside a program's anticipated logic---such as perceptual confusion between similar-looking objects---a fixed program may have limited means to revise its subsequent strategy in light of the failure. Our approach allows a program to perform multiple observations and actions per invocation, while the agent uses execution feedback at checkpoints or after failures to decide how to proceed. This aims to balance execution efficiency and adaptability, but model-call latency and discrete checkpoints still limit our response to rapidly changing contacts compared with the high-frequency visuomotor control targeted by VLA policies such as $\pi_0$~\citep{black2024pi0}. Future systems could combine high-level programmatic reasoning with faster low-level closed-loop control. Tools such as Jev~\citep{almeida2026jev}, designed for fast structured decisions, offer another direction to investigate, although their benefit for robot tasks remains to be established.

\section{Conclusion}
\label{sec:conclusion}

We present Agentic RSR, a Real-to-Sim-to-Real loop connecting reconstruction of a captured workspace, execution-grounded programmatic policy development, and real-robot execution. Robot-anchored registration establishes metric scale; DSH agents use reference-view depth and appearance feedback to guide scene reconstruction, while MuJoCo~\citep{todorov2012mujoco} checks task-relevant physical interactions. Policy development progresses from privileged information to visual observations and then randomized simulation. A program can observe and act repeatedly within one invocation, and the agent uses execution outcomes to decide how to continue or recover. During transfer, the policy and experience memory carry task knowledge into real-robot execution; the agent adapts its subsequent behavior to real observations, while robot-side safety mechanisms constrain its actions. The central contribution is a task-level loop in which scene reconstruction, executable policies, and real-world execution feedback inform one another.

%% file: section/06_ai_use.tex
\subsection*{AI use statement}

Generative AI tools, including coding agents, assisted with research and evaluation planning, implementation of parts of the Real-to-Sim and simulation-policy workflows, organization of experimental records and references, and manuscript polishing and typesetting. The role of AI agents within Agentic RSR is described separately in Section~\ref{sec:method}. The authors retain responsibility for the study design, verification of reported results and citations, and the final manuscript.

%% file: appendix/task_catalog.tex
\section{Task Catalog and ID Mapping}
\label{app:task-catalog}

The task identifiers below are used throughout the scene and policy evaluations.

\begin{table}[!ht]
\caption{Task IDs and descriptions for the nine Franka FR3 and nine Agilex Piper tasks.}
\label{tab:task-catalog}
\centering\footnotesize
\setlength{\tabcolsep}{4pt}
\begin{tabular}{@{}lp{0.38\linewidth}lp{0.38\linewidth}@{}}
\toprule
ID & Franka task & ID & Piper task \\
\midrule
F1 & Close the drawer & P1 & Close the cup lid \\
F2 & Put corn in the bucket & P2 & Put the brown block in the white-and-green storage box \\
F3 & Put the cup in the box and close the box & P3 & Put the emergency-stop button in the green storage box \\
F4 & Stack the snacks & P4 & Put the glue in the clear cup \\
F5 & Stack two bowls & P5 & Put the glue in the pen holder \\
F6 & Store fruit in the bamboo tray & P6 & Stack the brown blocks \\
F7 & Store fruit on the plate & P7 & Stack the cones \\
F8 & Store three tools & P8 & Stack the storage boxes \\
F9 & Store two beanbags & P9 & Take out the tissue \\
\bottomrule
\end{tabular}
\end{table}

%% file: appendix/stage1.tex
\input{appendix/stage1_data}

\section{Scene Reconstruction and Experiment Setup}
\label{app:scene-setup}
\label{app:stage1}

\subsection{Registration objective and settings}

We align the meter-scale robot surface to the reconstructed robot point cloud by estimating a similarity transform $T(\mathbf{x})=sR\mathbf{x}+\mathbf{t}$, where $s$ converts meters to VGGT units~\citep{wang2026vggtomega}. Candidate transforms are scored using projected robot masks, bidirectional nearest-neighbor distances, point-cloud coverage, and depth consistency. Let $\overline{I}$ and $\overline{R}$ be mean mask IoU and recall, and let $r$ be the Euclidean distance between the componentwise 5th and 95th percentiles of the observed robot point cloud. For each direction, $\rho$ is the 60th-percentile nearest-neighbor distance divided by $r$; $O$ is the smaller fraction of points whose nearest-neighbor distance is below $0.05r$. Let $d_{50}^{\mathrm{VGGT}}$ be the median across frames of the median absolute projected-depth residual on valid robot-mask/depth overlap. The dimensionless ranking loss is
\[
\begin{aligned}
L={}&0.50(1-\overline{I})+0.12(1-\overline{R})
 +0.14\min(5\rho_{m\to o},1)+0.08\min(5\rho_{o\to m},1)\\
 &+0.08(1-O)+0.08\min(d_{50}^{\mathrm{VGGT}}/r,1).
\end{aligned}
\]
When no valid depth overlap is available, the last term uses its maximum penalty. Table~\ref{tab:stage1-settings} summarizes the registration settings.

\begin{table}[!ht]
\caption{Settings for robot-based metric-scale registration.}
\label{tab:stage1-settings}
\centering\footnotesize
\begin{tabular}{@{}p{0.30\linewidth}p{0.63\linewidth}@{}}
\toprule
Setting & Value \\
\midrule
Sampled frames per capture & 32 \\
Rotation initialization & 128 candidates; refine the top 3 and select the lowest final loss \\
Initial alignment & Multi-scale Sim(3) ICP \\
Pose refinement & Two passes over translation, rotation, and scale, followed by coarse-to-fine rotation search \\
\bottomrule
\end{tabular}
\end{table}

\subsection{Registration summary}

Table~\ref{tab:stage1-summary} summarizes registration fit across nine captures per robot. $L_f$ is the final ranking loss; IoU is averaged across frames. The task-median projected-depth residual $d_{50}=10^3d_{50}^{\mathrm{VGGT}}/s$ is in millimeters, where $s$ converts meters to VGGT units~\citep{wang2026vggtomega}. These fit statistics are distinct from reference-pose error and downstream scene-quality metrics.

\begin{table}[!ht]
\caption{Robot-registration results across nine captures per robot.}
\label{tab:stage1-summary}
\centering\footnotesize
\begin{tabular}{@{}lcccccc@{}}
\toprule
Robot & $n$ & Mean $L_f$ & Median $L_f$ & $L_f$ range & Mean IoU & Task-median $d_{50}$ (mm) \\
\midrule
Franka & 9 & 0.11364 & 0.11221 & 0.10624--0.12115 & \textbf{0.8438} & 35.3 \\
Piper  & 9 & \textbf{0.10464} & \textbf{0.10373} & 0.09997--0.11216 & 0.8405 & \textbf{21.3} \\
\bottomrule
\end{tabular}
\end{table}

%% file: appendix/stage1_data.tex
\newcommand{\stageOneFrankaRows}{%
F1 & 0.10624 & 0.17494 & 0.8519 & 0.9763 & 35.3 & 0.5800 & 1 \\
F2 & 0.11385 & 0.28598 & 0.8558 & 0.9817 & 39.3 & 0.5355 & 1 \\
F3 & 0.11465 & 0.29848 & 0.8434 & 0.9624 & 27.7 & 0.5371 & 1 \\
F4 & 0.11196 & 0.26301 & 0.8397 & 0.9692 & 38.0 & 0.5312 & 1 \\
F5 & 0.11191 & 0.25840 & 0.8469 & 0.9757 & 33.3 & 0.5643 & 1 \\
F6 & 0.11221 & 0.24543 & 0.8433 & 0.9684 & 32.5 & 0.5342 & 1 \\
F7 & 0.12115 & 0.31231 & 0.8328 & 0.9701 & 37.3 & 0.5283 & 1 \\
F8 & 0.12021 & 0.26966 & 0.8371 & 0.9761 & 25.3 & 0.5293 & 1 \\
F9 & 0.11058 & 0.27465 & 0.8431 & 0.9789 & 41.4 & 0.5318 & 1 \\
}

\newcommand{\stageOnePiperRows}{%
P1 & 0.10251 & 0.13380 & 0.8395 & 0.9476 & 19.9 & 0.8238 & 1 \\
P2 & 0.10042 & 0.14351 & 0.8503 & 0.9567 & 22.1 & 0.8151 & 2 \\
P3 & 0.11216 & 0.18959 & 0.8413 & 0.9482 & 27.9 & 0.7961 & 3 \\
P4 & 0.10432 & 0.14583 & 0.8352 & 0.9587 & 20.2 & 0.7793 & 2 \\
P5 & 0.10807 & 0.14802 & 0.8392 & 0.9370 & 21.3 & 0.7900 & 1 \\
P6 & 0.10373 & 0.15450 & 0.8388 & 0.9491 & 19.1 & 0.8476 & 2 \\
P7 & 0.10927 & 0.15361 & 0.8308 & 0.9582 & 23.1 & 0.8517 & 2 \\
P8 & 0.09997 & 0.12980 & 0.8463 & 0.9523 & 21.3 & 0.8167 & 2 \\
P9 & 0.10130 & 0.14177 & 0.8434 & 0.9561 & 21.0 & 0.8599 & 2 \\
}

%% file: appendix/stage2.tex
\section{Codex Reconstruction Results}
\label{app:codex-history}
\label{app:quantitative-results}
\label{app:stage2}
\label{app:re3sim-audit}

This appendix preserves reconstruction results from the Codex implementation, separately from the DSH main experiments. The aggregate summary dated October 6, 2026 and the earlier per-scene ablation batch have distinct provenance; their statistics are not interchangeable.

\subsection{Latest Codex aggregate summary}

Table~\ref{tab:codex-models} reproduces the latest 18-scene Codex summaries. The GPT-6 Astra entry corresponds to the final delivered scenes with unified acceptance-threshold rechecking; its round limit is not specified in the supplied summary. GPT-5.6 Sol uses a ten-round limit. Both metric means use all 18 scenes, while first joint acceptance rounds average passing scenes only. The latest Sol metrics (0.20986~m, 13.3001, and 0.66912) differ from the earlier model-swap batch below (0.2119~m, 13.31, and 0.6702). No new per-scene records are available for the latest summaries, so the historical per-scene tables are not evidence for those latest Sol means.

\input{tab/codex_models}
\FloatBarrier

\subsection{Historical reconstruction and object geometry}

The earlier main configuration uses Codex agents with GPT-6 Astra at high reasoning effort. Its depth-only first-hit and selected delivery statistics below differ from the first joint acceptance statistic in Table~\ref{tab:codex-models}.

Across the 18 scenes, refinement reduces mean four-view Depth MAE from 0.9036~m to 0.1057~m and mean Lab $\Delta E_{76}$ from 24.84 to 11.04, while achieving a mean full-frame grayscale SSIM of 0.6990. All scenes reach mean Depth MAE $<0.15$~m by round 6; the mean first-hit and selected delivery rounds are 2.0 and 4.3. The depth threshold alone does not satisfy the complete acceptance criteria.

Across 42 measured objects, the scene-equal mean symmetric Chamfer distance is 0.0303~m, F1@1cm is 0.3655, and visible bounding-box center error is 0.0869~m. These results indicate that visual feedback improves appearance and depth consistency while recovering task-object arrangement. Historical per-scene results are reported below.

\subsection{Historical Codex component ablations}

We examine model choice, scalar metric feedback, and initial calibration on the same 18 scenes. The full system passes the visual check in all scenes; replacing the refinement agent with GPT-5.6 Sol at high reasoning effort reduces this to 10/18 and raises mean Depth MAE from 0.1057~m to 0.2119~m and Lab $\Delta E_{76}$ from 11.04 to 13.31. In this setting, GPT-6 Astra appears to use visual feedback more effectively, although this comparison does not establish a general advantage across spatial-reasoning tasks.

Without scalar metric feedback, eight scenes still reach the depth threshold at a mean first-hit round of 2.0, but only 6/18 pass the full visual check and mean Depth MAE rises to 0.3746~m. Images enable rapid adjustment, but explicit depth and appearance metrics better guide error prioritization and acceptance.

Without initial calibration, only 3/18 scenes pass. Of the four that reach the depth threshold, the mean first-hit round rises to 6.8. Lab $\Delta E_{76}$ remains close to the full system (11.99 versus 11.04), but Depth MAE increases to 0.2583~m. Thus, visually similar scenes can retain substantial position or scale errors; calibration provides a more reliable starting point. Table~\ref{tab:scene-ablation} summarizes these results, with per-scene data in Appendix~\ref{app:scene-refinement-ablations}.

\input{tab/scene_ablation}

\FloatBarrier

\subsection{Per-scene reconstruction comparison}

Table~\ref{tab:reconstruction-scenes} lists per-scene metrics for each method; task IDs are defined in Table~\ref{tab:task-catalog}. Lab $\Delta E_{76}$ is lower-is-better, grayscale SSIM is higher-is-better, and -- denotes an unreported metric. CD is symmetric Chamfer distance in meters, F1 uses a 1~cm threshold, and BBox is visible bounding-box center error in meters.

\begingroup
\tiny
\setlength{\tabcolsep}{1.6pt}
\begin{longtable}{@{}ll*{9}{r}@{}}
\caption{Historical per-scene reconstruction metrics for Franka and Piper; Agentic RSR refers to Codex GPT-6 Astra. Bold indicates the best comparable result among reported methods.}\label{tab:reconstruction-scenes}\\
\toprule
Method & Metric & 1 & 2 & 3 & 4 & 5 & 6 & 7 & 8 & 9 \\
\midrule
\endfirsthead
\multicolumn{11}{l}{\textit{Table \thetable\ continued}}\\
\toprule
Method & Metric & 1 & 2 & 3 & 4 & 5 & 6 & 7 & 8 & 9 \\
\midrule
\endhead
\multicolumn{11}{@{}l}{\textbf{Franka}} \\
\midrule
\input{tab/reconstruction_scene_matrix}
\end{longtable}
\endgroup

\subsection{Historical per-scene ablation details}
\label{app:scene-refinement-ablations}

In the earlier Codex ablation batch, we evaluate three variants on the same 18 reconstructed scenes, each with at most 10 refinement rounds. The model-swap variant uses GPT-5.6 Sol at high reasoning effort and passes the visual check on 10 scenes. The other two retain GPT-6 Astra at high reasoning effort while removing scalar metric feedback or initial calibration. Table~\ref{tab:scene-refinement-ablation-details} reports the delivered-scene evaluation for accepted scenes, the last valid evaluation for unsuccessful scenes, and the round-0 baseline when no valid refinement-round evaluation was produced.

\begingroup
\scriptsize
\setlength{\tabcolsep}{3pt}
\newcommand{\sceneAblationSol}{GPT-5.6 Sol}
\newcommand{\sceneAblationNoMetric}{No metric feedback}
\newcommand{\sceneAblationNoCalibration}{No initial calibration}
\begin{longtable}{@{}llccccc@{}}
\caption{Per-scene results for three scene-refinement ablations across 18 tasks.}\label{tab:scene-refinement-ablation-details}\\
\toprule
Variant & Task & Pass & Last round & $D$ (m) & Lab $\Delta E_{76}$ & SSIM \\
\midrule
\endfirsthead
\multicolumn{7}{l}{\textit{Table \thetable\ continued}}\\
\toprule
Variant & Task & Pass & Last round & $D$ (m) & Lab $\Delta E_{76}$ & SSIM \\
\midrule
\endhead
\input{tab/sol_scene_ablation_rows}
\input{tab/no_metric_feedback_ablation_rows}
\input{tab/no_initial_calibration_ablation_rows}
\end{longtable}
\endgroup

%% file: tab/codex_models.tex
\begin{table}[!ht]
\caption{Codex aggregate reconstruction results from the October 6, 2026 summary. Both configurations attempt 18 scenes. $R_{\mathrm{pass}}$ is the mean first joint acceptance round over passing scenes.}
\label{tab:codex-models}
\centering\scriptsize
\setlength{\tabcolsep}{2pt}
\begin{tabular*}{\linewidth}{@{\extracolsep{\fill}}lcccccc@{}}
\toprule
Model & $N_{\mathrm{metric}}$ & Pass & $R_{\mathrm{pass}}$ & MAE (m)$\downarrow$ & $\Delta E_{76}\downarrow$ & SSIM $\uparrow$ \\
\midrule
\input{tab/codex_model_rows}
\end{tabular*}
\end{table}

%% file: tab/codex_model_rows.tex
GPT-6 Astra & 18 & 18/18 & 3.80 & 0.10574 & 11.0359 & 0.69897 \\
GPT-5.6 Sol & 18 & 10/18 & 4.70 & 0.20986 & 13.3001 & 0.66912 \\
\bottomrule

%% file: tab/scene_ablation.tex
\begin{table}[!ht]
\caption{Historical Codex Real-to-Sim scene-refinement ablations across 18 scenes. $R_{0.15}$ is the mean first-hit round among scenes reaching the depth threshold.}
\label{tab:scene-ablation}
\centering\footnotesize
\setlength{\tabcolsep}{2pt}
\begin{tabular*}{\linewidth}{@{\extracolsep{\fill}}lccccc@{}}
\toprule
Variant & Pass & $R_{0.15}\downarrow$ & Depth MAE (m)$\downarrow$ & Lab $\downarrow$ & SSIM $\uparrow$ \\
\midrule
Codex GPT-6 Astra & \textbf{18/18} & \textbf{2.0} & \textbf{0.1057} & \textbf{11.04} & \textbf{0.6990} \\
GPT-5.6 Sol (high) & 10/18 & 3.9 & 0.2119 & 13.31 & 0.6702 \\
w/o metric feedback & 6/18 & 2.0 & 0.3746 & 15.12 & 0.6490 \\
w/o initial calibration & 3/18 & 6.8 & 0.2583 & 11.99 & 0.6597 \\
\bottomrule
\end{tabular*}
\end{table}

%% file: tab/reconstruction_scene_matrix.tex
Re$^3$Sim & Lab & 20.609 & 25.939 & 30.323 & 26.030 & 25.114 & 24.360 & 29.146 & 30.490 & 26.289 \\
 & SSIM & .5612 & .5334 & .4875 & .5280 & .5229 & .5028 & .4980 & .4708 & .5151 \\
RL-GSBridge & Lab & 25.854 & 29.812 & 31.825 & 29.304 & 25.436 & 27.795 & 31.520 & 31.740 & 27.745 \\
 & SSIM & .5507 & .5155 & .4297 & .4980 & .5465 & .5034 & .4644 & .4872 & .5296 \\
RoboSimGS & Lab & 20.818 & 24.948 & 28.042 & 24.814 & 23.658 & 23.916 & 26.386 & 28.163 & 24.992 \\
 & SSIM & .5758 & .5717 & .5233 & .5525 & .5589 & .5236 & .5511 & .5300 & .5405 \\
SimFoundry & Lab & 55.781 & 55.634 & 51.913 & 53.194 & 55.386 & 56.286 & 51.924 & 51.914 & 29.455 \\
 & SSIM & .4865 & .5303 & .5431 & .5370 & .5235 & .4887 & .5454 & .5515 & .6215 \\
 & Depth & .4808 & .4476 & -- & .8517 & .5711 & 1.5177 & -- & -- & .4680 \\
 & CD & .4756 & .2134 & .1003 & .1404 & .1612 & .2444 & .1160 & .0805 & .2485 \\
 & F1 & .0032 & .0000 & .0462 & .0027 & .0000 & .0000 & .0000 & .0578 & .0132 \\
 & BBox & .6709 & .2571 & .1631 & .1954 & .1922 & .2704 & .1539 & .1163 & .3892 \\
Agentic RSR (Codex) & Lab & \textbf{11.80} & \textbf{10.23} & \textbf{8.27} & \textbf{8.42} & \textbf{10.41} & \textbf{7.98} & \textbf{8.69} & \textbf{11.69} & \textbf{8.83} \\
 & SSIM & \textbf{.7116} & \textbf{.7649} & \textbf{.7748} & \textbf{.7183} & \textbf{.7468} & \textbf{.7585} & \textbf{.7632} & \textbf{.7263} & \textbf{.7263} \\
 & Depth & \textbf{.100} & \textbf{.104} & .100 & \textbf{.145} & \textbf{.080} & \textbf{.107} & .097 & .114 & \textbf{.100} \\
 & CD & \textbf{.0222} & \textbf{.0435} & \textbf{.0149} & \textbf{.0133} & \textbf{.0167} & \textbf{.0417} & \textbf{.0145} & \textbf{.0153} & \textbf{.0319} \\
 & F1 & \textbf{.2936} & \textbf{.0351} & \textbf{.4363} & \textbf{.4375} & \textbf{.3212} & \textbf{.1913} & \textbf{.3949} & \textbf{.3588} & \textbf{.1037} \\
 & BBox & \textbf{.0556} & \textbf{.0894} & \textbf{.0711} & \textbf{.0589} & \textbf{.0471} & \textbf{.1003} & \textbf{.0550} & \textbf{.0297} & \textbf{.0991} \\
\midrule
\multicolumn{11}{@{}l}{\textbf{Piper}} \\
\midrule
Re$^3$Sim & Lab & 15.487 & 14.695 & 15.855 & 15.386 & 14.669 & 13.772 & 14.391 & 14.995 & 13.304 \\
 & SSIM & .6420 & \textbf{.6501} & .6538 & .6516 & .6565 & \textbf{.6715} & .6565 & \textbf{.6319} & \textbf{.6688} \\
RL-GSBridge & Lab & 17.767 & 17.605 & 16.069 & 17.019 & 18.058 & 14.977 & 17.094 & 16.651 & 15.459 \\
 & SSIM & .5863 & .5993 & .6421 & .6168 & .5498 & .6440 & .6081 & .6116 & .6404 \\
RoboSimGS & Lab & 16.638 & 15.982 & 16.747 & 15.649 & 15.688 & 14.792 & 15.766 & 16.181 & 14.583 \\
 & SSIM & .6227 & .6341 & .6434 & .6501 & .6422 & .6486 & .6432 & .6228 & .6561 \\
SimFoundry & Lab & 73.283 & 70.678 & 74.325 & 72.335 & 73.594 & 65.375 & 74.235 & 69.383 & 73.186 \\
 & SSIM & .3446 & .3515 & .3344 & .3565 & .3387 & .3564 & .3326 & .3542 & .3296 \\
 & Depth & .1569 & .1731 & .1201 & .1291 & .1421 & .2595 & .1951 & .2144 & .1460 \\
 & CD & .1778 & .1665 & .0589 & .0545 & .1151 & .1069 & .3370 & .1835 & .1441 \\
 & F1 & .0000 & .0000 & .0000 & .0245 & .0000 & .0035 & .0000 & .0000 & .0000 \\
 & BBox & .1973 & .2058 & .2294 & .0821 & .1580 & .1455 & .3792 & \textbf{.2321} & \textbf{.2006} \\
Agentic RSR (Codex) & Lab & \textbf{12.58} & \textbf{14.50} & \textbf{11.61} & \textbf{12.40} & \textbf{11.84} & \textbf{11.25} & \textbf{13.07} & \textbf{12.51} & \textbf{12.57} \\
 & SSIM & \textbf{.6421} & .6358 & \textbf{.6805} & \textbf{.6583} & \textbf{.6697} & .6706 & \textbf{.6577} & .6156 & .6604 \\
 & Depth & \textbf{.084} & \textbf{.091} & \textbf{.093} & \textbf{.087} & \textbf{.124} & \textbf{.092} & \textbf{.148} & \textbf{.109} & \textbf{.129} \\
 & CD & \textbf{.0082} & \textbf{.0359} & \textbf{.0350} & \textbf{.0132} & \textbf{.0092} & \textbf{.0077} & \textbf{.0161} & \textbf{.0806} & \textbf{.1245} \\
 & F1 & \textbf{.6817} & \textbf{.3730} & \textbf{.0809} & \textbf{.5571} & \textbf{.7296} & \textbf{.7146} & \textbf{.2658} & \textbf{.4905} & \textbf{.1131} \\
 & BBox & \textbf{.0174} & \textbf{.0864} & \textbf{.0772} & \textbf{.0451} & \textbf{.0405} & \textbf{.0324} & \textbf{.0371} & .2856 & .3367 \\
\bottomrule

%% file: tab/sol_scene_ablation_rows.tex
\sceneAblationSol & F1 & No & 10 & 0.2110 & 10.79 & 0.6883 \\
\sceneAblationSol & F2 & Yes & 2 & 0.1198 & 10.75 & 0.7141 \\
\sceneAblationSol & F3 & No & 10 & 0.4005 & 15.46 & 0.6718 \\
\sceneAblationSol & F4 & No & 10 & 0.6543 & 27.77 & 0.5333 \\
\sceneAblationSol & F5 & Yes & 9 & 0.1389 & 8.89 & 0.7757 \\
\sceneAblationSol & F6 & No & 10 & 0.2516 & 11.46 & 0.7084 \\
\sceneAblationSol & F7 & No & 10 & 0.2916 & 8.34 & 0.7622 \\
\sceneAblationSol & F8 & Yes & 4 & 0.1479 & 7.62 & 0.7666 \\
\sceneAblationSol & F9 & No & 10 & 0.3517 & 12.23 & 0.6830 \\
\sceneAblationSol & P1 & Yes & 2 & 0.1209 & 14.55 & 0.6353 \\
\sceneAblationSol & P2 & Yes & 10 & 0.1498 & 12.92 & 0.6552 \\
\sceneAblationSol & P3 & Yes & 1 & 0.1085 & 14.31 & 0.6512 \\
\sceneAblationSol & P4 & Yes & 2 & 0.1410 & 14.66 & 0.6266 \\
\sceneAblationSol & P5 & Yes & 6 & 0.1394 & 12.77 & 0.6602 \\
\sceneAblationSol & P6 & No & 10 & 0.1767 & 16.72 & 0.5963 \\
\sceneAblationSol & P7 & No & 10 & 0.1431 & 14.00 & 0.6515 \\
\sceneAblationSol & P8 & Yes & 8 & 0.1262 & 13.29 & 0.6260 \\
\sceneAblationSol & P9 & Yes & 3 & 0.1415 & 13.07 & 0.6586 \\
\midrule

%% file: tab/no_metric_feedback_ablation_rows.tex
\sceneAblationNoMetric & F1 & No & 0 & 0.9390 & 29.92 & 0.4805 \\
\sceneAblationNoMetric & F2 & No & 10 & 0.1938 & 11.06 & 0.7450 \\
\sceneAblationNoMetric & F3 & No & 10 & 0.5095 & 26.04 & 0.5102 \\
\sceneAblationNoMetric & F4 & No & 10 & 0.2596 & 8.41 & 0.7656 \\
\sceneAblationNoMetric & F5 & No & 10 & 0.1616 & 7.04 & 0.7406 \\
\sceneAblationNoMetric & F6 & No & 10 & 0.2577 & 7.37 & 0.7313 \\
\sceneAblationNoMetric & F7 & No & 10 & 1.7704 & 14.70 & 0.7027 \\
\sceneAblationNoMetric & F8 & No & 0 & 0.9501 & 33.34 & 0.4854 \\
\sceneAblationNoMetric & F9 & No & 10 & 0.2661 & 10.52 & 0.7183 \\
\sceneAblationNoMetric & P1 & Yes & 2 & 0.1209 & 11.97 & 0.6708 \\
\sceneAblationNoMetric & P2 & Yes & 6 & 0.1493 & 12.62 & 0.6610 \\
\sceneAblationNoMetric & P3 & No & 10 & 0.2154 & 16.49 & 0.6338 \\
\sceneAblationNoMetric & P4 & Yes & 3 & 0.1348 & 11.60 & 0.6616 \\
\sceneAblationNoMetric & P5 & Yes & 2 & 0.1251 & 13.76 & 0.6334 \\
\sceneAblationNoMetric & P6 & No & 10 & 0.1488 & 16.54 & 0.6345 \\
\sceneAblationNoMetric & P7 & Yes & 5 & 0.1421 & 11.18 & 0.6650 \\
\sceneAblationNoMetric & P8 & No & 6 & 0.2630 & 17.30 & 0.5665 \\
\sceneAblationNoMetric & P9 & Yes & 1 & 0.1350 & 12.25 & 0.6762 \\
\midrule

%% file: tab/no_initial_calibration_ablation_rows.tex
\sceneAblationNoCalibration & F1 & Yes & 7 & 0.1388 & 9.54 & 0.5850 \\
\sceneAblationNoCalibration & F2 & No & 10 & 0.2074 & 10.32 & 0.7309 \\
\sceneAblationNoCalibration & F3 & No & 10 & 0.2830 & 10.71 & 0.7349 \\
\sceneAblationNoCalibration & F4 & No & 10 & 0.3699 & 10.36 & 0.7349 \\
\sceneAblationNoCalibration & F5 & Yes & 7 & 0.1477 & 9.01 & 0.7552 \\
\sceneAblationNoCalibration & F6 & No & 10 & 0.1599 & 8.65 & 0.6328 \\
\sceneAblationNoCalibration & F7 & No & 10 & 0.1483 & 8.54 & 0.7833 \\
\sceneAblationNoCalibration & F8 & No & 10 & 0.2331 & 6.83 & 0.7956 \\
\sceneAblationNoCalibration & F9 & No & 10 & 0.2354 & 8.35 & 0.7550 \\
\sceneAblationNoCalibration & P1 & No & 10 & 0.2061 & 15.24 & 0.5556 \\
\sceneAblationNoCalibration & P2 & Yes & 4 & 0.1467 & 13.41 & 0.6021 \\
\sceneAblationNoCalibration & P3 & No & 10 & 0.1719 & 13.48 & 0.6318 \\
\sceneAblationNoCalibration & P4 & No & 9 & 0.3461 & 15.16 & 0.6151 \\
\sceneAblationNoCalibration & P5 & No & 10 & 0.2426 & 15.07 & 0.5786 \\
\sceneAblationNoCalibration & P6 & No & 10 & 0.1725 & 13.65 & 0.6270 \\
\sceneAblationNoCalibration & P7 & No & 0 & 0.9969 & 18.21 & 0.5913 \\
\sceneAblationNoCalibration & P8 & No & 10 & 0.2573 & 15.05 & 0.5488 \\
\sceneAblationNoCalibration & P9 & No & 10 & 0.1863 & 14.18 & 0.6165 \\
\bottomrule

%% file: appendix/agent_prompt_templates.tex
\section{Agent Prompt Templates}
\label{app:prompts}

This section reproduces the task-independent prompt templates, rather than one
task's fully instantiated conversation. Braced fields are runtime substitutions;
local paths, private run locations, and identifying metadata have been removed.
Images and public execution reports supplied alongside a prompt are described in
the templates but are not reproduced here. The Real-to-Sim templates retain their original semantics for the DSH implementation; simulation-policy templates retain their existing Codex workflow. The historical physical-check template is preserved in the source at \path{appendix/prompts/codex/stage3_physics_prompt.md}; its session identity names Codex instead of DSH.

\newcommand{\prompttemplate}[1]{\VerbatimInput[frame=single,framerule=0.4pt,framesep=5pt,fontsize=\scriptsize,breaklines,breakanywhere]{#1}}

\subsection{Real-to-Sim}
\paragraph{Stage 1: robot mask query.}
The only Stage 1 prompt reproduced here is the robot-mask query.
\prompttemplate{appendix/prompts/stage1_robot_prompt.txt}

\paragraph{Stage 2: visual planning.}
The planner receives four reference RGB views with the following template.
\prompttemplate{appendix/prompts/stage2_plan_prompt.md}

\paragraph{Stage 2: iterative scene refinement.}
The refinement agent also receives the four current renders, the approved plan,
and scalar feedback. The no-metric-feedback and uncalibrated ablations modify
parts of this prompt and should not be confused with the default condition.
\prompttemplate{appendix/prompts/stage2_refinement_prompt.md}

\paragraph{Stage 3: physical-role classification.}
\prompttemplate{appendix/prompts/stage3_semantics_prompt.md}

\paragraph{Stage 3: image-only physical prior.}
This turn is issued only when the compiled scene contains a free-joint object.
\prompttemplate{appendix/prompts/stage3_visual_force_prior_prompt.md}

\paragraph{Stage 3: physical plausibility and repair.}
This prompt is sent after scene compilation; a failed independent check may
append a subsequent repair-feedback turn in the same conversation.
\prompttemplate{appendix/prompts/stage3_physics_prompt.md}

\paragraph{Stage 3: independent-check feedback.}
If the candidate fails verification, the same conversation receives this
repair turn with the check results substituted at runtime.
\prompttemplate{appendix/prompts/stage3_repair_feedback.txt}

\subsection{SIM Policy Iteration}
The three stages are \texttt{privileged\_exact}, \texttt{visual\_exact}, and
\texttt{visual\_randomized}, in that order. They use \emph{one} authoring
template with a stage-specific access note and optional preceding-stage policy
and experience; there are not three independent fixed system prompts.

\paragraph{Shared three-stage authoring template.}
\prompttemplate{appendix/prompts/sim_task_prompt.txt}

\paragraph{Practice feedback.}
During authoring, a single practice CLI command and its public result are
returned to the same conversation before policy submission.
\prompttemplate{appendix/prompts/sim_practice_feedback.txt}

\paragraph{Explicit execution request.}
The agent must request execution after proposing a policy.
\prompttemplate{appendix/prompts/sim_execution_request.txt}

\paragraph{Within-stage continuation and revision.}
Runtime results and the public action log are substituted into this template;
the wrist and third-person images may accompany the turn.
\prompttemplate{appendix/prompts/sim_continuation_prompt.txt}

\paragraph{Experience handoff between stages.}
After a completed stage, the agent summarizes public evidence for the next
stage's fresh conversation.
\prompttemplate{appendix/prompts/sim_experience_prompt.txt}